\documentclass[10pt,twocolumn,letterpaper]{article}

\usepackage{cvpr}              % To produce the CAMERA-READY version
\definecolor{cvprblue}{rgb}{0.21,0.49,0.74}
\usepackage[pagebackref,breaklinks,colorlinks,allcolors=cvprblue]{hyperref}
\usepackage{multirow}
\usepackage{makecell}
\usepackage{tabularx}
\usepackage{booktabs}
\usepackage{siunitx}
\usepackage{float}
\usepackage[accsupp]{axessibility}
\usepackage[table,xcdraw]{xcolor}

\definecolor{bestblue}{HTML}{4A86E8}    % neutral top1
\definecolor{secondblue}{HTML}{9CC3FF}  % neutral top2
\definecolor{bestred}{HTML}{E06666}     % emotional/efficiency top1
\definecolor{secondred}{HTML}{F4CCCC}   % emotional/efficiency top2

\usepackage{tikz}
\newcommand{\bestbluebox}{\tikz[baseline=-0.5ex]\draw[fill=bestblue,draw=bestblue] (0,0) rectangle (0.4em,0.4em);}
\newcommand{\secondbluebox}{\tikz[baseline=-0.5ex]\draw[fill=secondblue,draw=secondblue] (0,0) rectangle (0.4em,0.4em);}
\newcommand{\bestredbox}{\tikz[baseline=-0.5ex]\draw[fill=bestred,draw=bestred] (0,0) rectangle (0.4em,0.4em);}
\newcommand{\secondredbox}{\tikz[baseline=-0.5ex]\draw[fill=secondred,draw=secondred] (0,0) rectangle (0.4em,0.4em);}

\def\paperID{3762} % *** Enter the Paper ID here
\def\confName{CVPR}
\def\confYear{2026}

\title{EmoTaG: Emotion-Aware Talking Head Synthesis on Gaussian Splatting \\ with Few-Shot Personalization}

\author{Haolan Xu$^{1}$, Keli Cheng$^{2}$, Lei Wang$^{2}$, Ning Bi$^{2}$, Xiaoming Liu$^{1,3}$\\
$^{1}$Michigan State University, East Lansing, MI 48824\\
$^{2}$Qualcomm Technologies Inc., San Diego, CA 92121\\
$^{3}$University of North Carolina at Chapel Hill, Chapel Hill, NC 27599\\
{\tt\small xuhaola2@msu.edu, \{kelic, wlei, nbi\}@qti.qualcomm.com, liuxm@cs.unc.edu} 
}
\begin{document}

\twocolumn[{%
\maketitle
\begin{figure}[H]
\hsize=\textwidth % cvpr
\centering
\includegraphics[width=0.92\textwidth]{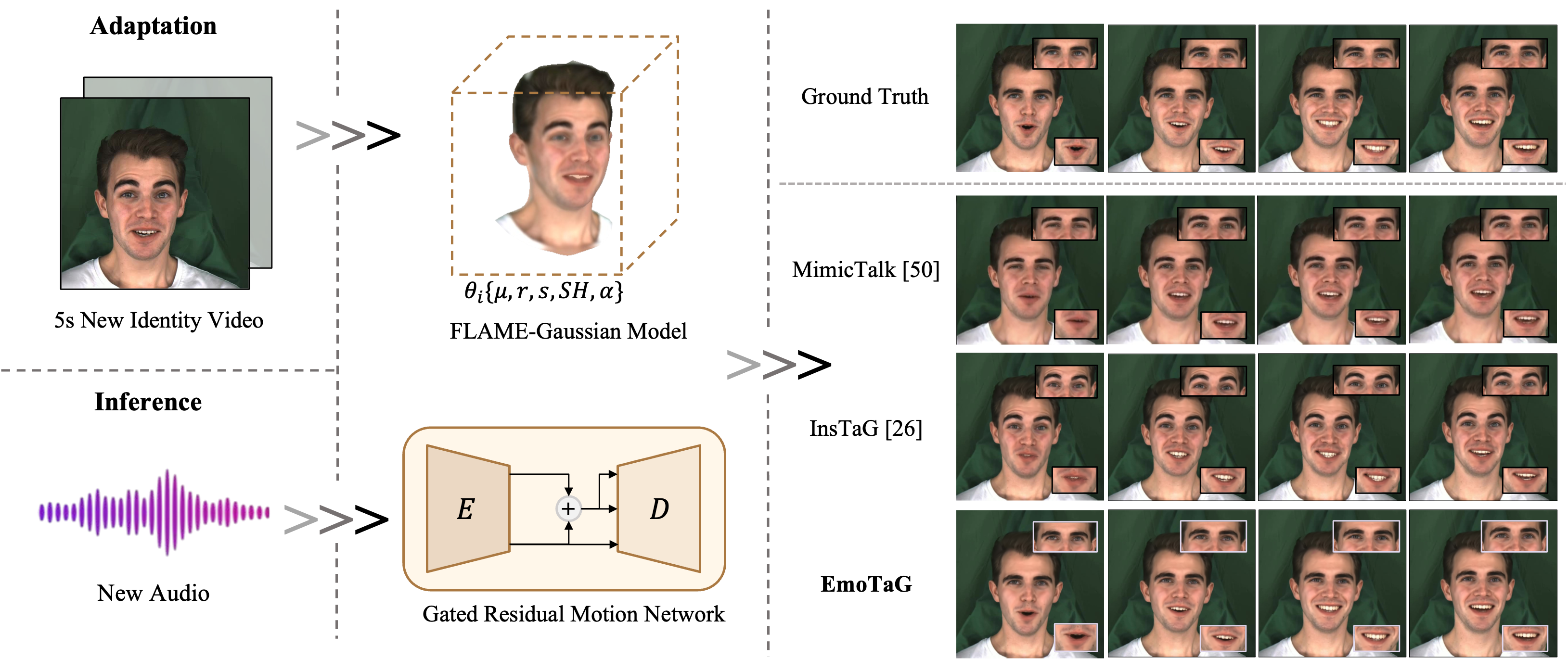}
\caption{EmoTaG generates expressive and synchronized 3D talking heads from only 5-second new identity videos.
Built upon a FLAME-Gaussian model~\cite{qian2024gaussianavatars} and a Gated Residual Motion Network, our method achieves better emotional expressiveness, lip synchronization, visual realism, and motion stability compared to state-of-the-art approaches~\cite{ye2024mimictalk, li2025instag}.}
\label{teaser}
\end{figure}
}]

\begin{abstract}
Audio-driven 3D talking head synthesis has advanced rapidly with Neural Radiance Fields (NeRF) and 3D Gaussian Splatting (3DGS). By leveraging rich pre-trained priors, few-shot methods enable instant personalization from just a few seconds of video. However, under expressive facial motion, existing few-shot approaches often suffer from geometric instability and audio-emotion mismatch, highlighting the need for more effective emotion-aware motion modeling. In this work, we present EmoTaG, a few-shot emotion-aware 3D talking head synthesis framework built on the Pretrain-and-Adapt paradigm. Our key insight is to reformulate motion prediction in a structured FLAME parameter space rather than directly deforming 3D Gaussians, thereby introducing explicit geometric priors that improve motion stability. Building upon this, we propose a Gated Residual Motion Network (GRMN), which captures emotional prosody from audio while supplementing head pose and upper-face cues absent from audio, enabling expressive and coherent motion generation. Extensive experiments demonstrate that EmoTaG achieves state-of-the-art performance in emotional expressiveness, lip synchronization, visual realism, and motion stability. \href{https://emotag26.github.io/}{Project page}.
\end{abstract}    
\section{Introduction}
\label{sec:intro}
Audio-driven talking head synthesis, which aims to generate realistic facial animations synchronized with audio, has become a key technology in digital humans and virtual reality~\cite{gafni2021dynamic, tan2024style2talker, tan2024say, chen2025echomimic, zhang2024tamm, zhang2025unleashing, yin2025trim, kumar2025charm3r, zhu2026fusionagent, zhu2026can, zhu2025quality, zielonka2023instant}. The advent of Neural Radiance Fields (NeRF)~\cite{mildenhall2021nerf} and 3D Gaussian Splatting (3DGS)~\cite{kerbl20233d} has significantly advanced 3D talking head synthesis~\cite{li2023efficient, ye2023geneface, li2024talkinggaussian, cho2024gaussiantalker, li2025instag, hou2024compose, liu2022blind}. But current 3D person-specific models~\cite{li2023efficient, li2024talkinggaussian, zhang2025fate, li2025rgbavatar} still rely on minute-level monocular videos and subject-wise optimization, which impedes rapid deployment and scalable personalization. 

\begin{figure}[t] 
\centering 
\includegraphics[width=\linewidth]{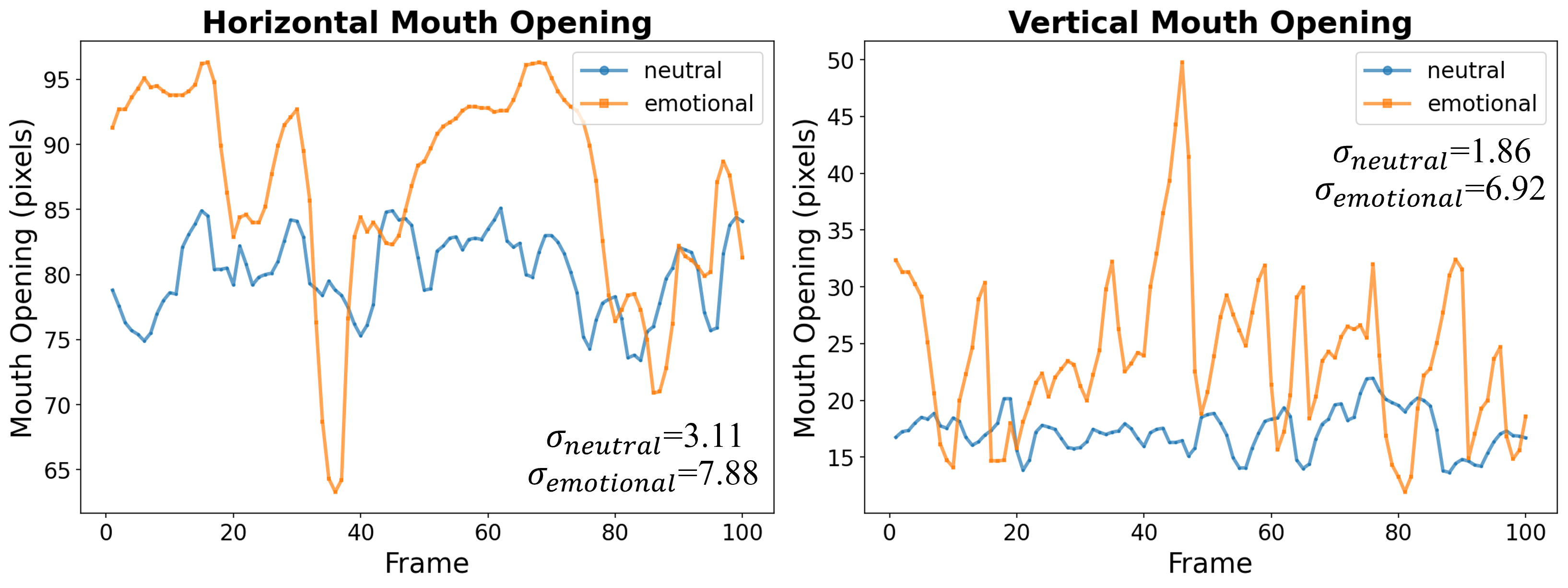} 
\vspace{-15pt}
\caption{
\textbf{Neutral vs emotional articulation complexity.} 
We compare horizontal and vertical mouth opening trajectories measured from lip landmarks. 
Emotional audio shows stronger temporal fluctuation and larger standard deviations (\(\sigma_{\text{emotional}}=7.88, 6.92\)) than neutral audio (\(\sigma_{\text{neutral}}=3.11, 1.86\)), 
revealing more complex articulation patterns.
}
\vspace{-10pt}
\label{fig:mouth_open} 
\end{figure}

A promising direction is the Pretrain-and-Adapt (PAA) paradigm that learns a universal audio-motion prior and adapts it to new identities using a few seconds of video. However, existing PAA-based pipelines~\cite{li2025instag, nie2025few} primarily focus on neutral speech scenarios with limited ability to capture emotion-driven facial motion, a central component of real-world human communication. Empirical evidence in Fig.~\ref{fig:mouth_open} further shows that emotional audio exhibits more complex motion dynamics than neutral audio, highlighting the need to model emotion-driven prosodic cues beyond phonemes. This raises a core question: \textit{Can few-shot 3D talking head synthesis move beyond neutral speech to support emotion-aware animation for expressive speech?}

To answer this question, we propose Emotion-Aware Talking Head Synthesis on Gaussian Splatting \textbf{(EmoTaG)}, a framework for emotion-aware and structurally consistent 3D talking head synthesis with few-shot personalization, as shown in Fig.~\ref{teaser}.
Specifically, we adopt FLAME~\cite{li2017learning} as an explicit geometric prior and predict its expression and jaw parameters. These predicted parameters deform the FLAME mesh and drive rigged 3D Gaussians through a FLAME-Gaussian formulation~\cite{qian2024gaussianavatars} for stable facial motion. Intra-oral regions are further refined using mouth Gaussians selected by lip landmarks, which capture fine-grained articulations such as teeth and tongue movements.
Building on this, we propose a Gated Residual Motion Network (GRMN) composed of an Identity-Conditioned Encoder and an Expert Motion Decoder. The encoder integrates prosody-aware audio features and upper-face Action Unit cues to capture emotion-related dynamics, while incorporating an identity embedding through AdaIN-based modulation~\cite{huang2017arbitrary, karras2019style} for personalized feature adaptation. The decoder contains three cooperative branches: a base branch that captures phonetic audio-motion mapping, a residual branch that models emotion-related and identity-specific variations, and a gate branch that adaptively combines the base and residual outputs to produce stable and expressive facial motion. To enhance emotional sensitivity, we further introduce Semantic Emotion Guidance, which distills the emotion distribution and emotion intensity score from a pretrained emotion recognizer named DeepFace~\cite{serengil2024lightface} into the GRMN’s residual and gate branches, enabling emotion-aware motion learning without manual annotations.

Our main contributions are summarized as follows:
\begin{itemize}
\item We propose \textbf{EmoTaG}, a framework for emotion-aware and stable few-shot 3D talking head synthesis that adapts to a new identity using only 5 seconds of video.
\item We design a Gated Residual Motion Network that separates phonetic articulation from emotion-related motion and adaptively fuses them through a gating mechanism. We further introduce Semantic Emotion Guidance, which distills semantic emotion knowledge from a pretrained emotion recognizer to enable emotion-aware motion learning without manual annotations.
\item Extensive experiments show that EmoTaG achieves state-of-the-art performance in emotional expressiveness, lip synchronization, visual realism, and motion stability across diverse evaluation scenarios.
\end{itemize}

\begin{figure*}[ht]
    \centering
    \includegraphics[width=1\textwidth]{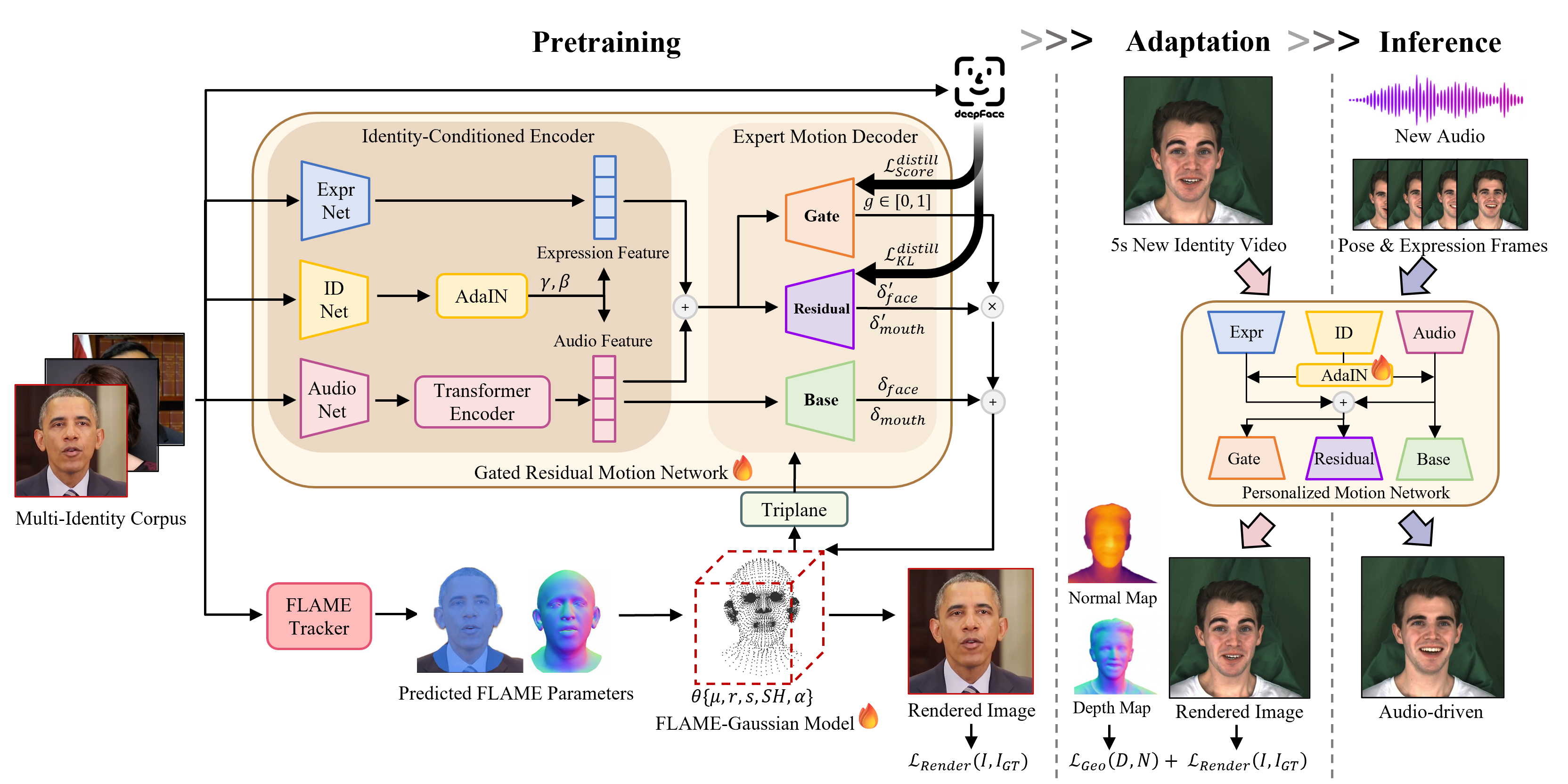}
    \vspace{-15pt}
    \caption{
\textbf{Overview of EmoTaG.} 
For pretraining, our Gated Residual Motion Network learns a universal motion prior from a multi-identity corpus. This network comprises an Identity-Conditioned Encoder for integrating audio, expression, and identity through AdaIN-based modulation, followed by an Expert Motion Decoder that leverages emotion-distilled supervision to train three cooperative branches (Base, Residual, Gate). During adaptation, the Gated Residual Motion Network is efficiently adapted to a new identity from 5-second video via only tuning the AdaIN modulation parameters. At inference, the adapted model produces expressive, high-fidelity 3D facial animation driven by new audio with head pose and upper-face cues.
}
    \vspace{-10pt}
    \label{pipeline}
\end{figure*}

\section{Related Works}
\label{sec:formatting}

%-------------------------------------------------------------------------
\subsection{Audio-Driven 3D Talking Head Synthesis}
With the emergence of NeRF~\cite{mildenhall2021nerf} and 3D Gaussian Splatting (3DGS)~\cite{kerbl20233d}, 3D talking head synthesis has become a promising direction. Existing 3D person-specific methods ~\cite{li2024talkinggaussian, zhang2025fate, li2025rgbavatar, hou2022face, hou2021towards} train a model for each identity from scratch, achieving high-fidelity results but requiring several minutes of video. To reduce this cost, one-shot methods~\cite{ye2024real3d, hu2025ggtalker, tan2025fixtalk} adapt pretrained priors from a single portrait, but their limited identity information often yields less personalized and unstable results. Alternatively, few-shot Pretrain-and-Adapt methods such as InsTaG~\cite{li2025instag} and FIAG~\cite{nie2025few} adapt a universal motion prior learned from a multi-identity corpus to new identities using only a few seconds of video, enabling efficient and realistic personalization. However, these few-shot methods face two key limitations: the lack of explicit emotional modeling, leading to less expressive facial animations; and unconstrained 3DGS deformation, which causes geometric instability under diverse and exaggerated expressions. Our work addresses both limitations by introducing emotion-aware motion learning with a strong structural prior.

%-------------------------------------------------------------------------
\subsection{Emotional Talking Head Synthesis}
Natural talking head synthesis requires not only accurate lip synchronization but also emotionally expressive facial dynamics. Recent 2D generative approaches~\cite{tan2023emmn, tan2024edtalk, zakharov2019few, xu2024hallo, tan2024flowvqtalker, tan2025disentangle, tan2025edtalk++} produce vivid expressions, but the absence of explicit 3D structural priors makes it difficult to preserve consistent geometry and realistic deformation, particularly under large head poses or highly expressive facial motions. In the 3D domain, EMOTE~\cite{danvevcek2023emotional} and EmoVOCA~\cite{nocentini2025emovoca} rely on 3D face datasets with manual emotion annotations to train FLAME-based models. However, such annotation-driven pipelines are restricted by coarse discrete emotion categories and annotator subjectivity, preventing them from learning fine-grained emotional dynamics directly from audio. EmoTalk3D~\cite{he2024emotalk3d} achieves high-fidelity results with 3D Gaussian Splatting, but its reliance on person-specific optimization significantly limits scalability for rapid personalization. To our knowledge, no methods simultaneously achieve few-shot adaptation, 3D consistency, and emotion-aware modeling from audio, as achieved by EmoTaG.

\section{Method}
As illustrated in Fig.~\ref{pipeline}, EmoTaG consists of two major components:
(a) FLAME-Gaussian Model (Sec.~\ref{sec:flame_gaussian}), which provides a robust structural 3D prior; and 
(b) Gated Residual Motion Network (GRMN), which predicts dynamic facial motion. 
GRMN contains two modules: an Identity-Conditioned Encoder (Sec.~\ref{sec:encoder}), which integrates multimodal driving cues through AdaIN-based modulation, and an Expert Motion Decoder (Sec.~\ref{sec:decoder}) that employs three complementary motion branches. We further incorporate Semantic Emotion Guidance (Sec.~\ref{sec:distill}), which distills emotion distributions and intensity scores from a pretrained emotion recognizer~\cite{serengil2024lightface} to guide the residual and gate branches in the decoder without manual annotations.

%-------------------------------------------------------------------------
\subsection{Preliminaries}

\textbf{FLAME Parametric Model.}
FLAME~\cite{li2017learning} is a statistical head model based on linear blend skinning, which represents facial geometry with a compact set of interpretable parameters. 
Given shape parameters $\boldsymbol{\beta}$, expression parameters $\boldsymbol{\Psi}$, and pose parameters $\boldsymbol{\Theta}$, FLAME outputs a mesh:
\begin{equation}
M(\boldsymbol{\beta}, \boldsymbol{\Psi}, \boldsymbol{\Theta}) \in \mathbb{R}^{3 \times N},
\label{eq:flame-mesh}
\end{equation}
where $N$ denotes the number of vertices. 
In our work, $\boldsymbol{\beta}$ is fixed to preserve identity, while the motion prediction task is defined as regressing $\boldsymbol{\Psi}$ and the jaw-related pose subset $\boldsymbol{\Theta}_{\text{jaw}} \in \mathbb{R}^{3}$. This formulation provides a strong structural prior that ensures stability across different identities. 

\noindent \textbf{3D Gaussian Splatting.}
3DGS~\cite{kerbl20233d} represents a scene by anisotropic Gaussians:
\begin{equation}
\mathcal{G}=\{g_i\}_{i=1}^{K},\quad
g_i=(\boldsymbol{\mu}_i,\boldsymbol{r}_i,\boldsymbol{s}_i,\alpha_i,\mathbf{SH}_i),
\label{eq:gauss-def}
\end{equation}
where $\boldsymbol{\mu}_i\in\mathbb{R}^3$ is the Gaussian center, $\boldsymbol{r}_i$ and $\boldsymbol{s}_i$ denote its rotation and scale vectors used to construct the covariance, $\alpha_i$ is opacity, and $\mathbf{SH}_i$ are spherical-harmonic coefficients for view-dependent color. This explicit point-based representation enables high-quality real-time rendering. We initialize the 3D Gaussians $\mathcal{G}$ by uniformly sampling an equal number of points per FLAME mesh triangle \(T=(\mathbf{v}_1,\mathbf{v}_2,\mathbf{v}_3)\). Each Gaussian center is computed via barycentric interpolation:
\begin{equation}
\boldsymbol{\mu}_i = \sum_{j=1}^{3} w_{ij}\mathbf{v}_j,\quad 
\sum_{j=1}^{3}w_{ij}=1,\; w_{ij}\ge0.
\end{equation}
The barycentric weights bind each Gaussian to its parent triangle, allowing consistent deformation by updating $\mathbf{v}_j$ during FLAME mesh deformation.

The intra-oral Gaussian subset $\mathcal{G}_{\text{mouth}}\!\subset\!\mathcal{G}$ is derived by identifying a mouth region \(\Omega_{\text{mouth}}\) on an augmented FLAME mesh completed with intra-oral geometry following \cite{qian2024gaussianavatars}. We lift 2D mouth landmarks to 3D via barycentric interpolation on this augmented mesh, map them to the nearest vertices, and expand them via geometry-constrained region growing to obtain a contiguous intra-oral surface. Gaussians with centers fall within \(\Omega_{\text{mouth}}\) are selected to form \(\mathcal{G}_{\text{mouth}}\), providing an anatomically coherent initialization for modeling intra-oral dynamics.

%-------------------------------------------------------------------------
\subsection{Gated Residual Motion Network}

\subsubsection{Identity-Conditioned Encoder}
\label{sec:encoder}
The GRMN fuses audio, expression, and identity into a unified motion representation. Specifically, we encode:
(i) an audio feature $\mathbf{A}$ to capture phonetic content and prosodic rhythm, 
(ii) an expression feature $\mathbf{E}$ to supply the frame-wise upper-face expression information (\textit{e.g.}, brow raise, eye squeeze) absent in $\mathbf{A}$, and
(iii) an identity feature $\mathbf{s}$ to preserve identity-specific motion style.

We obtain the audio feature $\mathbf{A}$ by first feeding the raw waveform through a pretrained Wav2Vec 2.0 encoder~\cite{baevski2020wav2vec} to extract frame-level speech embeddings. Following AD-NeRF~\cite{guo_ad-nerf_2021}, we refine these embeddings using a temporal 1D CNN, and then apply an MLP to project them into the input space of a 4-layer Transformer encoder, which further enriches the representation with long-range temporal and prosodic cues.
For each video frame, we compute Action Unit (AU) parameters using OpenFace~\cite{baltruvsaitis2015cross,8373812}, providing complementary upper-face expression cues. The AU parameters are subsequently mapped into a compact embedding by an MLP-based AU encoder, yielding the expression feature $\mathbf{E}$. 
To model personalized motion characteristics, we compute the identity feature $\mathbf{s}$ by selecting the top-50 neutral frames ranked by DeepFace~\cite{serengil2024lightface} and averaging their AdaFace~\cite{kim2022adaface} features, resulting in a compact identity descriptor.
The identity feature $\mathbf{s}$ is then used to modulate the other two feature streams via feature-wise normalization using Adaptive Instance Normalization (AdaIN)~\cite{huang2017arbitrary, karras2019style}.
Formally, given an input feature $\mathbf{F} \in \{\mathbf{A}, \mathbf{E}\}$ and the identity feature $\mathbf{s}$, the modulation parameters $(\boldsymbol{\gamma}, \boldsymbol{\beta})$ are predicted using an MLP:
\begin{equation}
\gamma, \beta = \mathrm{MLP}(\mathbf{s}),
\end{equation}
and the modulated feature $\tilde{\mathbf{F}}$ is computed as:
\begin{equation}
\tilde{\mathbf{F}} = \gamma \cdot \mathrm{InstanceNorm}(\mathbf{F}) + \beta.
\label{eq:adain}
\end{equation}
This process injects personalized motion characteristics into both audio and expression features before they are fused for motion prediction. After modulation, the fused feature is passed through the Expert Motion Decoder, which contains three cooperative branches.

\subsubsection{Expert Motion Decoder}
\label{sec:decoder}
\noindent \textbf{Base branch.}
This branch models identity-agnostic audio-motion mapping, focusing on neutral speech articulation disentangled from emotion and identity. Although the input audio features are identity-modulated via AdaIN, this branch is supervised to capture shared phoneme-driven deformation patterns across multiple identities, predicting base deformation parameters ${\boldsymbol{\delta}}_{\text{b}}=(\boldsymbol{\delta}_{\text{face}}, \boldsymbol{\delta}_{\text{mouth}})$ that represent neutral lower-face and mouth motion. Here, \(\boldsymbol{\delta}_{\text{face}}\) denotes the FLAME expression and jaw pose offsets for global facial motion, while \(\boldsymbol{\delta}_{\text{mouth}}\) represents intra-oral Gaussian deformation offsets for fine-grained mouth articulation.

\noindent \textbf{Residual branch.}
To capture emotion-driven deviations, we introduce a residual branch consisting of an MLP-based EMO encoder-decoder module. The EMO encoder takes the modulated and concatenated audio-expression features to produce an emotion latent $\mathbf{z}_e$, while the EMO decoder maps $\mathbf{z}_e$ to motion residuals ${\boldsymbol{\delta}}_{\text{r}}=(\boldsymbol{\delta}'_{\text{face}}, \boldsymbol{\delta}'_{\text{mouth}})$. This branch leverages upper-face cues to enhance emotion-related facial motion, providing complementary signals that are absent in audio. This design increases responsiveness to emotion-driven variations. Its training is guided by the Semantic Emotion Guidance described in Sec.~\ref{sec:distill}.

\noindent \textbf{Gate branch.}
Because emotional intensity varies across frames, the gating unit predicts a scalar \(g \in [0,1]\) that adaptively fuses neutral and emotion-related motion:
\begin{equation}
{\boldsymbol{\delta}}
= {\boldsymbol{\delta}}_{\text{b}} + g \cdot {\boldsymbol{\delta}}_{\text{r}}.
\label{eq:grmn-fuse}
\end{equation}
The gate dynamically regulates the contribution of emotion-related motion, preventing over-exaggeration and preserving structural stability. The fused motion representation \(\hat{\boldsymbol{\delta}}\) is used to drive the FLAME-Gaussian model, producing geometrically consistent and emotion-aware facial motion.

\subsubsection{FLAME-Gaussian Motion Mapping}
\label{sec:flame_gaussian}
Given the predicted FLAME expression and jaw pose parameters $(\boldsymbol{\Psi}, \boldsymbol{\Theta}_{\text{jaw}})$, we propagate mesh deformation to the 3D Gaussians through a rigged mapping scheme~\cite{qian2024gaussianavatars}. Each 3D Gaussian \(g_i \in \mathcal{G}\) is bound to a parent FLAME mesh triangle \(j\), and is defined in the triangle’s local coordinate frame while being transformed into the global space as the triangle deforms. Let \(\mathbf{R}^j\), \(\mathbf{C}^j\), and \(k^j\) denote the rotation matrix, center, and isotropic scale of the FLAME mesh triangle \(j\). For each Gaussian \(g_i\), we denote its local attributes in the triangle's coordinate frame with subscript \(l\), including position \(\boldsymbol{\mu}_l\), rotation \(\boldsymbol{r}_l\), scale \(\boldsymbol{s}_l\), opacity \(\alpha_l\), and spherical harmonics \(\mathbf{SH}_l\). The global attributes of Gaussian \(g_i\) are then computed via:
\begin{equation}
\mathcal{G}_i =
\begin{cases}
\boldsymbol{\mu}_i = k^j \mathbf{R}^j \boldsymbol{\mu}_l + \mathbf{C}^j,\\[2pt]
\boldsymbol{r}_i = \mathbf{R}^j \boldsymbol{r}_l,\\[2pt]
\boldsymbol{s}_i = k^j \boldsymbol{s}_l,\\[2pt]
\alpha_i = \alpha_l,\\[2pt]
\mathbf{SH}_i = \mathbf{SH}_l.
\end{cases}
\label{eq:rigged-gaussian}
\end{equation}
This mapping preserves the structural consistency of the FLAME mesh while allowing each Gaussian to move coherently with its corresponding surface patch. 

For Gaussians inside the intra-oral region $\mathcal{G}_{\text{mouth}}$, we further apply the network-predicted residual offsets $(\Delta\boldsymbol{\mu}, \Delta\boldsymbol{r}, \Delta\boldsymbol{s})$ to enhance fine-grained articulations:
\begin{equation}
\boldsymbol{\mu}^* = \boldsymbol{\mu} + \Delta\boldsymbol{\mu}, \quad
\boldsymbol{r}^* = \boldsymbol{r} + \Delta\boldsymbol{r}, \quad
\boldsymbol{s}^* = \boldsymbol{s} + \Delta\boldsymbol{s}.
\label{eq:residual-gaussian}
\end{equation}
Here, $\alpha$ and $\mathbf{SH}$ remain fixed since they primarily encode static appearance. This hybrid deformation preserves structural motion from FLAME while introducing intra-oral details through learned Gaussian refinements.

% SEG Figure
\begin{figure}[t] 
\centering 
\includegraphics[width=\linewidth]{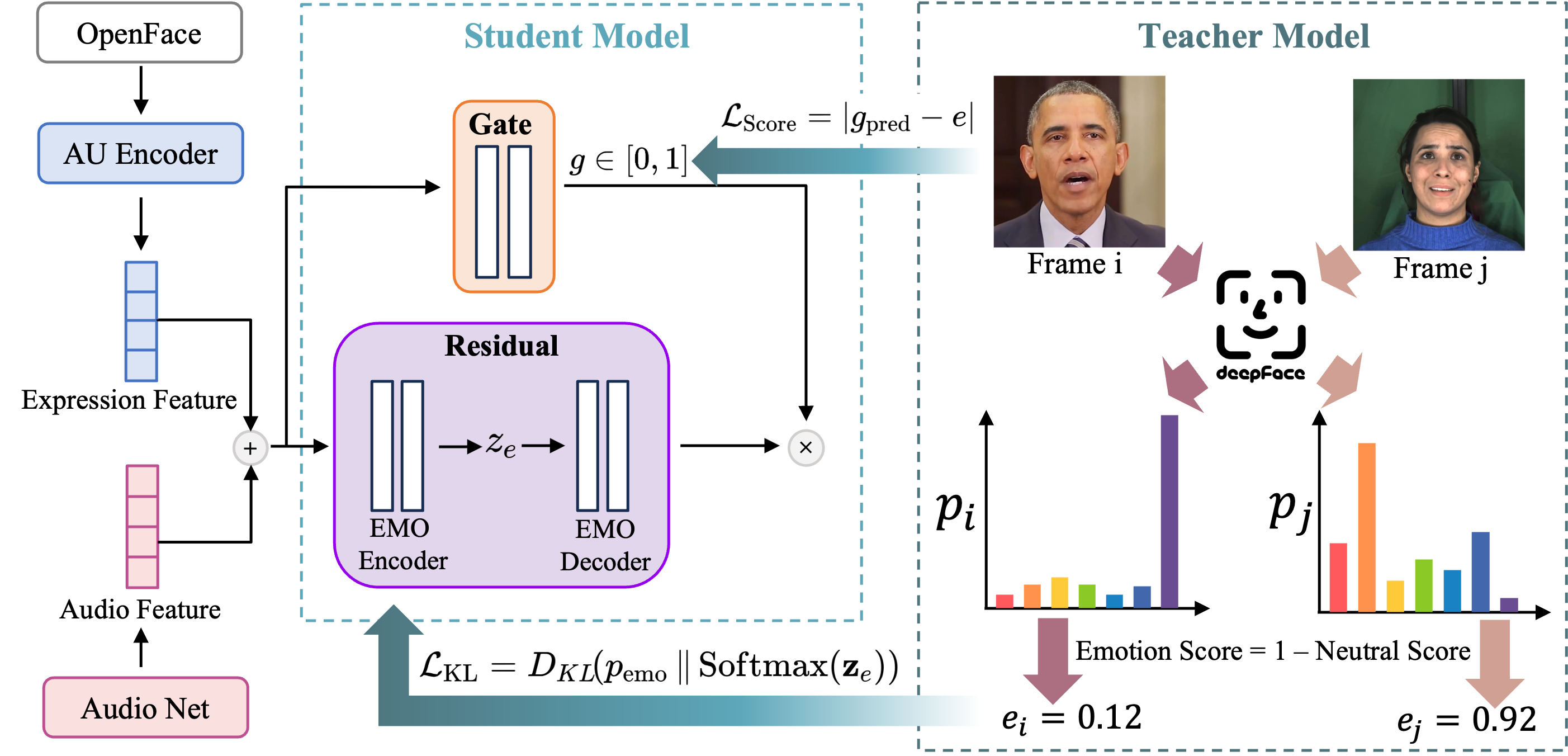}
\vspace{-15pt}
\caption{ \textbf{Overview of Semantic Emotion Guidance.} DeepFace provides both a categorical emotion distribution and a scalar emotion score to guide GRMN’s residual and gate branches. } 
\vspace{-10pt}
\label{fig:seg} 
\end{figure}

\subsubsection{Semantic Emotion Guidance}
\label{sec:distill}
While GRMN supports structured motion decomposition, we incorporate explicit emotional supervision to enhance its sensitivity to emotion-driven variations. Given the scarcity and bias of manual annotations, we adopt a teacher-student distillation framework termed Semantic Emotion Guidance, as illustrated in Fig.~\ref{fig:seg}. An off-the-shelf emotion recognizer, DeepFace~\cite{serengil2024lightface}, serves as the teacher and provides two supervisory signals for each training frame:
(i) a normalized emotion distribution \(p_{\text{emo}}\) representing the probabilities of seven basic emotions, and 
(ii) a scalar emotion score \(e\) that measures the intensity of the current expression, defined as:
\begin{equation}
e = 1 - p_{\text{emo}}(\text{neutral}).
\label{eq:emotion-intensity}
\end{equation}

These teacher signals guide two submodules in GRMN: the residual branch captures emotion information by aligning its latent representation \(\mathbf{z}_e\) with the teacher’s emotion distribution \(p_{\text{emo}}\), while the gate branch learns emotion intensity by matching the scalar \(e\). This dual guidance enables GRMN to model both the type and intensity of emotions, yielding expressive yet controllable motion generation.

% Main result figure
\begin{figure*}[ht]
    \centering
    \includegraphics[width=1\textwidth]{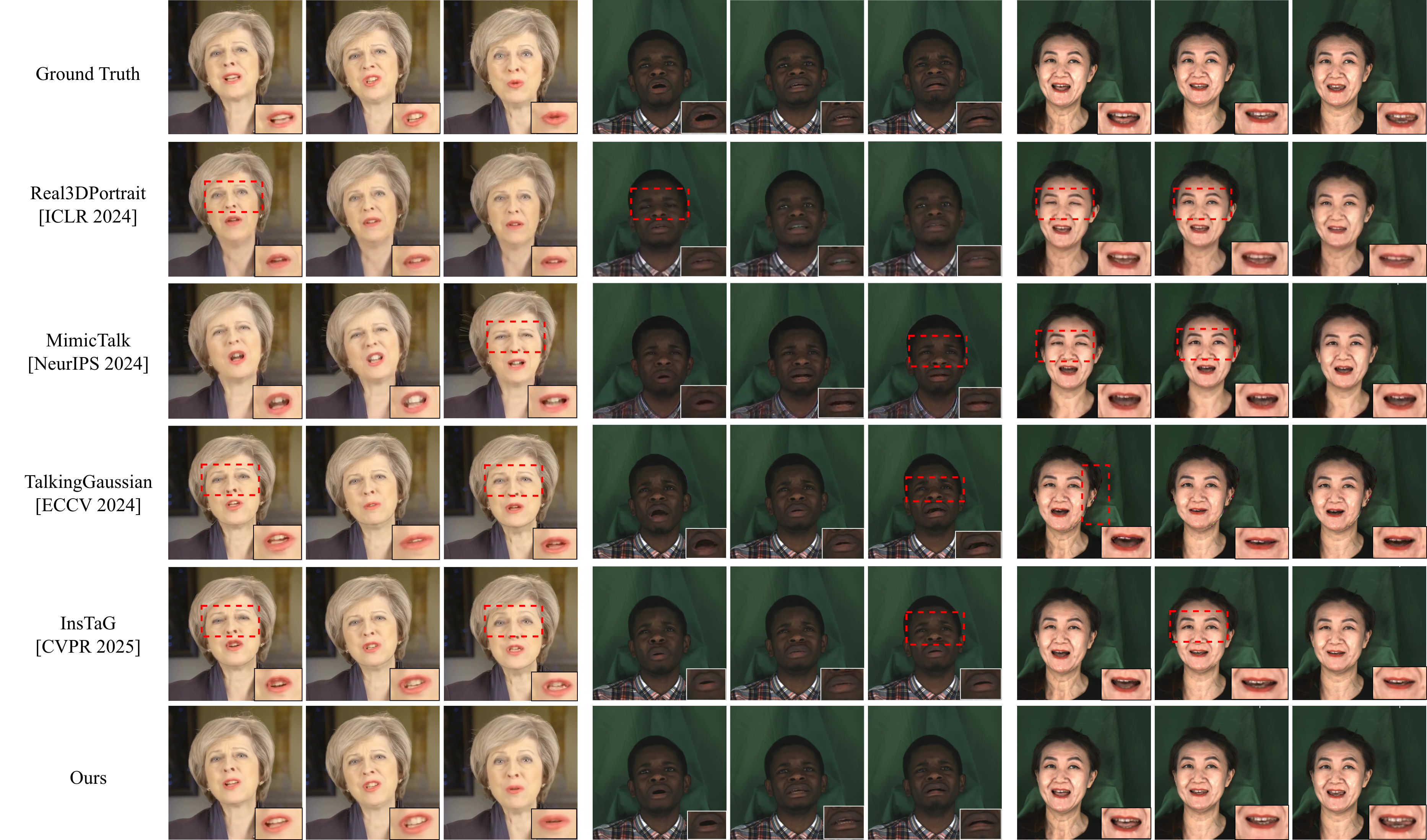}
    \vspace{-10pt}
    \caption{
\textbf{Qualitative comparison on \textit{self-reconstruction}.} EmoTaG generates more expressive, temporally coherent, and well-synchronized talking heads than previous methods across both neutral and emotional test cases. It preserves accurate mouth articulation and natural upper-face motion even under challenging expressions. The red rectangles highlight regions where previous methods produce distorted and inaccurate facial motions. We strongly recommend watching the supplementary video for dynamic comparison.}
    \vspace{-10pt}
    \label{fig:Qual_Results}
\end{figure*}

\subsection{Pretrain \& Adaptation Framework}
During pretraining, the Gated Residual Motion Network (GRMN) learns a universal motion prior from a multi-identity corpus. During adaptation, we achieve efficient personalization by freezing the pretrained GRMN and only fine-tuning the AdaIN modulation parameters, which enables EmoTaG to capture identity-specific dynamics from just a 5-second video. Once the GRMN has been adapted, we perform inference using this personalized motion network. Because audio carries very limited information about upper-face expression or head pose, we additionally provide a set of pose \& expression frames as auxiliary inputs during inference, following the previous works~\cite{ye2024real3d,ye2024mimictalk,li2025instag}. These frames are processed by the FLAME Tracker to get head pose, and by the Expression Network to obtain upper-face cues. With new audio as the driving signal and the pose \& expression cues offering complementary conditioning, the adapted GRMN produces expressive and coherent facial motions that naturally correspond to the new audio.

\subsection{Training Objectives}

\label{sec:training}
Our full objective combines photometric, emotional, and geometric supervision:
\begin{equation}
\mathcal{L} = 
\mathcal{L}_{\text{Render}} +
\mathcal{L}_{\text{KL}} +
\mathcal{L}_{\text{Score}} +
\mathcal{L}_{\text{Geo}},
\end{equation}
where the first three losses are used in both pretraining and adaptation, while $\mathcal{L}_{\text{Geo}}$ is applied only during adaptation.

\noindent \textbf{Rendering loss.} 
Following 3D Gaussian Splatting~\cite{kerbl20233d}, the rendering loss ensures both pixel and structural fidelity:
\begin{equation}
\mathcal{L}_{\text{Render}} =
\mathcal{L}_{1}(I, I_{GT}) +
\lambda_{\text{D-SSIM}}\!\cdot\!(1 - \mathrm{SSIM}(I, I_{GT})),
\end{equation}
where $\mathcal{L}_{1}$ enforces color consistency, while SSIM~\cite{wang2004image} preserves perceptual structure.

\noindent \textbf{Emotion distillation losses.}  
To distill explicit emotion cues from the DeepFace~\cite{serengil2024lightface} teacher, the residual branch aligns its latent emotion representation $\mathbf{z}_e$ with the teacher-provided emotion distribution $p_{\text{emo}}$:
\begin{equation}
\mathcal{L}_{\text{KL}} = D_{KL}\!\left(p_{\text{emo}} \,\|\, \mathrm{Softmax}(\mathbf{z}_e)\right),
\end{equation}
while the gate branch regresses the teacher’s scalar emotion score $e$:
\begin{equation}
\mathcal{L}_{\text{Score}} = |g_{\text{pred}} - e|.
\end{equation}

\noindent \textbf{Geometric loss.}  
Following Gaussianpro~\cite{cheng2024gaussianpro}, we compute the 2D depth $D$ and normal $N$ maps from the rendered image $I$ during adaptation. These maps are compared against pseudo-ground-truths $D_{GT}$ and $N_{GT}$ from Sapiens~\cite{khirodkar2024sapiens}:
\begin{equation}
\mathcal{L}_{\text{Geo}} =
\mathcal{L}_{D}(D, D_{GT}) +
\mathcal{L}_{N}(N, N_{GT}).
\end{equation}
This auxiliary loss mitigates overfitting under limited visual supervision and preserves physically consistent facial geometry across frames.

% Main Table
\begin{table*}[t]
\centering
\small
\resizebox{\textwidth}{!}{
\setlength{\tabcolsep}{3pt}
\begin{tabular}{l|ccccccc|ccccccc|cc}
\toprule
\multirow{2}{*}{Method}
& \multicolumn{7}{c|}{\textbf{Neutral Set}} 
& \multicolumn{7}{c|}{\textbf{Emotional Set}} 
& \multicolumn{2}{c}{\textbf{Efficiency}} \\
\cline{2-17}
& PSNR$\uparrow$ & LPIPS$\downarrow$ & SSIM$\uparrow$
& LMD$\downarrow$ & \multicolumn{2}{c}{\rule{0pt}{2.4ex}AUE-(L/U)$\downarrow$} & Sync-C$\uparrow$
& PSNR$\uparrow$ & LPIPS$\downarrow$ & SSIM$\uparrow$
& LMD$\downarrow$ & \multicolumn{2}{c}{\rule{0pt}{2.4ex}AUE-(L/U)$\downarrow$} & Sync-C$\uparrow$
& Train$\downarrow$ & FPS$\uparrow$ \\
\midrule
Real3DPortrait~\cite{ye2024real3d}
& 25.41 & 0.053 & 0.827
& 3.415 & 1.143 & 1.084 & \cellcolor{bestblue}6.719
& 25.16 & 0.063 & 0.815
& 3.642 & 1.183 & 1.149 & \cellcolor{bestred}6.583
& -- & 8.9 \\
ER-NeRF~\cite{li2023efficient}
& 28.21 & 0.038 & 0.847
& 3.549 & 1.314 & 0.466 & 3.142
& 27.31 & 0.054 & 0.822
& 4.215 & 1.541 & 0.802 & 2.714
& 2 h & 33.2 \\
TalkingGaussian~\cite{li2024talkinggaussian}
& 28.43 & 0.034 & 0.844
& 3.582 & 1.167 & \cellcolor{secondblue}0.401 & 3.631
& \cellcolor{secondred}27.84 & 0.042 & 0.836
& 4.392 & 1.213 & \cellcolor{secondred}0.618 & 3.059
& 27 min & \cellcolor{bestred}118.4 \\
GeneFace++~\cite{ye2023geneface++}
& 27.58 & 0.044 & 0.835
& 3.521 & 1.212 & 0.647 & 3.737
& 27.16 & 0.047 & 0.824
& 3.721 & 1.253 & 0.961 & 3.588
& 7 h & 19.3 \\
MimicTalk~\cite{ye2024mimictalk}
& 25.26 & 0.071 & 0.857
& 3.478 & 0.964 & 0.781 & \cellcolor{secondblue}6.341
& 25.13 & 0.079 & 0.842
& 3.577 & \cellcolor{secondred}0.961 & 0.913 & 6.113
& 17 min & 8.6 \\
InsTaG~\cite{li2025instag}
& \cellcolor{secondblue}28.92 & \cellcolor{secondblue}0.029 & \cellcolor{secondblue}0.866
& \cellcolor{secondblue}3.145 & \cellcolor{secondblue}0.921 & 0.407 & 5.329
& 27.82 & \cellcolor{secondred}0.040 & \cellcolor{secondred}0.851
& \cellcolor{secondred}3.428 & 0.995 & 0.651 & 4.828
& \cellcolor{secondred}13 min & \cellcolor{secondred}82.5 \\
\textbf{EmoTaG (Ours)}
& \cellcolor{bestblue}30.02 & \cellcolor{bestblue}0.019 & \cellcolor{bestblue}0.883
& \cellcolor{bestblue}2.221 & \cellcolor{bestblue}0.685 & \cellcolor{bestblue}0.210 & 6.212
& \cellcolor{bestred}29.95 & \cellcolor{bestred}0.022 & \cellcolor{bestred}0.877
& \cellcolor{bestred}2.456 & \cellcolor{bestred}0.702 & \cellcolor{bestred}0.236 & \cellcolor{secondred}6.147
& \cellcolor{bestred}11 min & 76.4 \\
\bottomrule
\end{tabular}}
\vspace{-5pt}
\caption{\textbf{Quantitative comparison on \textit{self-reconstruction}} of neutral and emotional talking videos with 5s training data. \protect\bestbluebox{} and \protect\secondbluebox{} indicate the 1st and 2nd best results on the neutral set, while \protect\bestredbox{} and \protect\secondredbox{} indicate the 1st and 2nd best results on the emotional set. Efficiency columns (Train, FPS) are ranked using the red palette.}
\vspace{-10pt}
\label{tab:main}
\end{table*}

% Emotion Intensity
\begin{table}[t]
\centering
\small
\resizebox{\linewidth}{!}{
\setlength{\tabcolsep}{3pt}
\begin{tabular}{l|ccccc}
\toprule
Methods & PSNR$\uparrow$ & LMD$\downarrow$ & \multicolumn{2}{c}{AUE-(L/U)$\downarrow$} & Sync-C$\uparrow$ \\
\midrule
\multicolumn{6}{c}{\textbf{Level-1 (Weaker Intensity)}} \\
\midrule
Real3DPortrait~\cite{ye2024real3d} & 25.27 & 3.581 & 1.148 & 1.094 & \textbf{6.593} \\
MimicTalk~\cite{ye2024mimictalk}     & 25.18 & 3.504 & \underline{0.946} & 0.895 & 6.129 \\
InsTaG~\cite{li2025instag}              & \underline{28.03} &\underline{3.352} & 0.968 & \underline{0.613} & 5.014 \\
\rowcolor{gray!20} \textbf{EmoTaG (Ours)}                  & \textbf{30.01} & \textbf{2.367} & \textbf{0.697} & \textbf{0.221} & \underline{6.154} \\
\midrule
\multicolumn{6}{c}{\textbf{Level-3 (Stronger Intensity)}} \\
\midrule
Real3DPortrait~\cite{ye2024real3d} & 25.11 & 3.742 & 1.215 & 1.238 & \textbf{6.585} \\
MimicTalk~\cite{ye2024mimictalk}      & 25.05 & 3.637 & \underline{0.981} & 0.933 & 6.118 \\
InsTaG~\cite{li2025instag}              & \underline{27.56} & \underline{3.559} & 1.144 & \underline{0.704} & 4.637 \\
\rowcolor{gray!20} \textbf{EmoTaG (Ours)}                  & \textbf{29.92} & \textbf{2.522} & \textbf{0.721} & \textbf{0.244} & \underline{6.126} \\
\bottomrule
\end{tabular}
}
\vspace{-5pt}
\caption{
\textbf{Quantitative comparison on \textit{emotion-intensity}.} We adapt models on Level-2 (medium) and test on Level-1 (weaker) and Level-3 (stronger) separately.
[Key: \textbf{Best}, \underline{Second Best}]}
\vspace{-10pt}
\label{tab:emotion_intensity}
\end{table}

\section{Experiment}

\textbf{Datasets.}
We train EmoTaG on the HDTF dataset~\cite{zhang2021flow}, using 70 videos (90-240 seconds each) from 70 identities to learn identity-agnostic audio-motion priors. For evaluation, we construct two test sets: a neutral set comprising 10 public videos of 10 identities from~\cite{li2023efficient, ye2023geneface}, and an emotional set derived from MEAD~\cite{wang2020mead}, covering 5 emotion categories-\textit{happy}, \textit{sad}, \textit{surprised}, \textit{angry}, and \textit{fear}-with 3 intensity levels and 2 identities per emotion. All videos are face-centered cropped and resized to $512{\times}512$ at 25 FPS.

\noindent \textbf{Baselines.}
We compare EmoTaG with representative 3D talking-head baselines across different training settings.
Train-from-scratch methods include ER-NeRF~\cite{li2023efficient} and TalkingGaussian~\cite{li2024talkinggaussian}, which require full retraining per identity.
In the one-shot setting, we evaluate Real3DPortrait~\cite{ye2024real3d}, which enables rapid identity initialization from a single reference image.
For few-shot adaptation, we compare against GeneFace++~\cite{ye2023geneface++}, MimicTalk~\cite{ye2024mimictalk}, and InsTaG~\cite{li2025instag}.
All baselines are reproduced using their official implementations, with Wav2Vec~2.0~\cite{baevski2020wav2vec} as the unified audio feature extractor for fair comparison.

\noindent \textbf{Metrics.}
We evaluate EmoTaG in terms of rendering fidelity, motion accuracy, lip synchronization, and efficiency.
Image quality is assessed using PSNR, SSIM~\cite{wang2004image}, and LPIPS~\cite{zhang2018unreasonable} between rendered and ground-truth frames.
Facial-motion accuracy is measured by Landmark Distance (LMD) and Action Unit Error for the lower face and upper face (AUE-L/U)~\cite{ekman1978facial, li2024talkinggaussian}.
Audio-visual synchronization is quantified with SyncNet-based Confidence (Sync-C) and Error (Sync-E)~\cite{chung2016lip, chung2016out}.
Efficiency is reported in terms of training time and inference FPS.

\noindent \textbf{Implementation Details.}
We employ VHAP~\cite{qian2024vhap} as our FLAME tracker to obtain the FLAME shape parameters, non-jaw head pose, and camera parameters. The 3D Gaussian field is initialized with 60K Gaussians uniformly sampled from the FLAME mesh. Training follows a two-stage curriculum. For each identity, the first 1,000 iterations optimize static appearance only, after which Gaussian parameters and the Gated Residual Motion Network are jointly optimized. Pretraining and adaptation are performed for $250$K and $20$K iterations, respectively, using AdamW~\cite{loshchilov2017decoupled} with learning rates of $5{\times}10^{-3}$ and $5{\times}10^{-4}$, correspondingly. Loss weights are set to $\lambda_{\text{D-SSIM}}{=}2{\times}10^{-1}$, $\lambda_{D}{=}1{\times}10^{-2}$, and $\lambda_{N}{=}1{\times}10^{-3}$. All experiments are conducted on a single NVIDIA RTX~A6000 GPU. In our method, audio encoding is performed once per sequence ($\sim$25 ms), and each frame requires $\sim$6 ms for GRMN inference and $\sim$7 ms for 3DGS rendering. Additional implementation details are provided in the supplementary material.

%-------------------------------------------------------------------------
\subsection{Evaluation}
\textbf{Comparison Settings.} 
We evaluate EmoTaG under three settings: (1) \textit{Self-reconstruction}, where each model is adapted on a 5-second video and evaluated on another clip (3-10 seconds) of the same emotion and identity, measuring few-shot rendering quality, motion accuracy, and lip synchronization.
(2) \textit{Emotion-intensity}, where models adapted on Level-2 are tested on Level-1 and Level-3 audio of the same emotion and identity to assess sensitivity to intensity variation.
(3) \textit{Out-of-distribution (OOD) audio-driven}, which evaluates generalization to unseen audio while keeping the adapted identity fixed. We consider (a) cross-identity—speech from a different speaker in the same language, and (b) cross-language—speech in a different language from the adaptation clip. For each case, we sample three audio clips from public videos.
Following the prior work~\cite{li2025instag}, we use pose \& expression cues from the test clip for \textit{self-reconstruction} and \textit{emotion-intensity} settings, and from the adaptation clip for \textit{OOD audio-driven}.

% Cross domain
\begin{table}[t]
\centering
\small
\resizebox{\linewidth}{!}{
\setlength{\tabcolsep}{3pt}
\begin{tabular}{l|cc|cc}
\toprule
\multirow{2}{*}{Methods} &
\multicolumn{2}{c|}{\textit{Cross Identity}} &
\multicolumn{2}{c}{\textit{Cross Language}} \\
& \makecell{Sync-E $\downarrow$} 
& \makecell{Sync-C $\uparrow$} 
& \makecell{Sync-E $\downarrow$} 
& \makecell{Sync-C $\uparrow$} \\
\midrule
ER-NeRF~\cite{li2023efficient}          & 11.732 & 2.833 & 11.931 & 2.542 \\
GeneFace++~\cite{ye2023geneface++}      & 10.515 & 3.497 &  11.052  & 3.107 \\
TalkingGaussian~\cite{li2024talkinggaussian} & 10.407 & 3.207 & 10.938 & 2.941 \\
InsTaG~\cite{li2025instag}        & \underline{9.921} & \underline{4.722} & \underline{10.033} & \underline{4.391} \\
\midrule
\textbf{EmoTaG (Ours)}         & \textbf{9.133} & \textbf{5.814} & \textbf{9.662} & \textbf{5.432} \\
\bottomrule
\end{tabular}
}
\vspace{-5pt}
\caption{
\textbf{Quantitative results of lip synchronization on \textit{OOD audio-driven}.} 
We evaluate two challenging scenarios: \textit{cross-identity} and \textit{cross-language}. 
[Key: \textbf{Best}, \underline{Second Best}]}
\vspace{-10pt}
\label{tab:cross_domain}
\end{table}

\noindent \textbf{Quantitative Results.}
Tab.~\ref{tab:main} reports the \textit{self-reconstruction} results.
Across both neutral and emotional test sets, EmoTaG achieves the best overall performance, with the highest PSNR/SSIM and lowest LPIPS, indicating superior appearance fidelity and geometric stability. Its lowest LMD and AUE-(L/U) further reflect accurate lip articulation and natural upper-face motion. Although EmoTaG is not the best on Sync-C, it remains close to the two large-scale pretrained models~\cite{ye2024real3d, ye2024mimictalk} and substantially outperforms them on the remaining metrics. EmoTaG adapts in 11 minutes and runs in real time at 76.4~FPS.
Tab.~\ref{tab:emotion_intensity} evaluates \textit{emotion-intensity} sensitivity.
When adapted on Level-2 and tested on Level-1/Level-3, EmoTaG obtains the lowest LMD and AUE-(L/U), demonstrating precise modulation across intensity levels while maintaining competitive Sync-C and the highest visual quality. Notably, its performance gains over InsTaG are larger at higher intensity, highlighting greater robustness to intense emotions.
In the \textit{out-of-distribution} evaluation (Tab.~\ref{tab:cross_domain}), EmoTaG achieves the lowest Sync-E and highest Sync-C in both \textit{cross-identity} and \textit{cross-language} settings, confirming better audio-motion generalization and lip synchronization under unseen speakers and languages.

\noindent \textbf{Qualitative Comparison.}
Fig.~\ref{fig:Qual_Results} presents qualitative comparisons under the \textit{self-reconstruction} setting. 
Compared with existing approaches, EmoTaG produces more expressive results with stable identity preservation. 
TalkingGaussian~\cite{li2024talkinggaussian} exhibits visible artifacts and mouth-audio misalignment, while InsTaG~\cite{li2025instag} improves lip synchronization but lacks natural emotional variation. 
MimicTalk~\cite{ye2024mimictalk} and Real3DPortrait~\cite{ye2024real3d} achieve better alignment due to large-scale pretraining yet still generate rigid and over-smoothed expressions. 
Using strong structural priors and emotion-aware motion learning, EmoTaG reconstructs photorealistic details and consistent facial dynamics in both neutral and emotional tests. 
We strongly recommend viewing the supplementary video for dynamic comparison.

% User Study
\begin{table}[t]
\centering
\small
\resizebox{\linewidth}{!}{
\setlength{\tabcolsep}{3pt}
\begin{tabular}{l|ccc}
\toprule
Method & Emo Expr & Lip Sync & Visual Realism \\
\midrule
Real3DPortrait~\cite{ye2024real3d} & 2.20 & 3.50 & 3.20 \\
TalkingGaussian~\cite{li2024talkinggaussian}             & 3.30 & 2.40 & 3.00 \\
MimicTalk~\cite{ye2024mimictalk}             & 3.50 & \underline{4.10} & 3.70 \\
InsTaG~\cite{li2025instag}     & \underline{3.80} & 3.90 & \underline{4.20} \\
\midrule
\textbf{EmoTaG (Ours)}                & \textbf{4.50} & \textbf{4.70} & \textbf{4.60} \\
\bottomrule
\end{tabular}
}
\vspace{-5pt}
\caption{
\textbf{User study on emotional expressiveness, lip synchronization, and visual realism.} 
Ratings range from 1 to 5, where higher indicates better performance. 
[Key: \textbf{Best}, \underline{Second Best}]}
\vspace{-10pt}
\label{tab:user-study}
\end{table}

\noindent \textbf{User Study.}
To assess perceptual quality, we conduct a user study comparing EmoTaG against representative baselines.
Twenty participants rate anonymized results from the \textit{self-reconstruction} setting on emotional expressiveness, lip synchronization, and visual realism (1-5 Likert, higher is better).
As shown in Tab.~\ref{tab:user-study}, EmoTaG achieves the best overall scores, notably on emotional expressiveness, demonstrating stronger emotion-aware and coherent facial motion.

%-------------------------------------------------------------------------
\subsection{Ablation Study}
\begin{figure}[t] 
\centering 
\includegraphics[width=\linewidth]{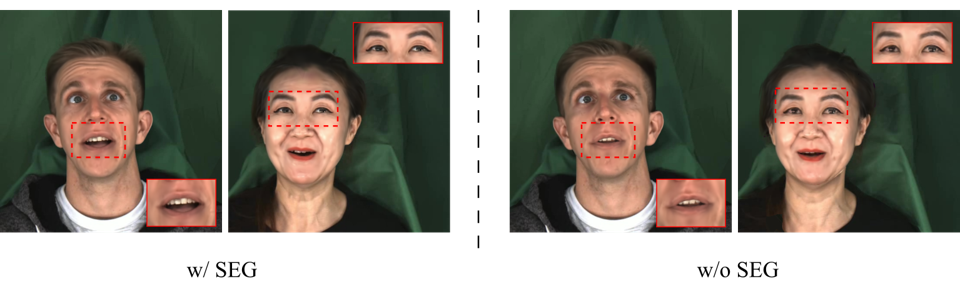} 
\vspace{-20pt}
\caption{
\textbf{Effect of Semantic Emotion Guidance (SEG).} 
SEG enhances upper-face expressiveness and audio-emotion coherence compared to the model without it.} 
\vspace{-10pt}
\label{fig:ablation} 
\end{figure}

To evaluate the contribution of each component, we ablate by removing key modules of EmoTaG. Quantitative results are listed in Tab.~\ref{tab:ablation}, and qualitative examples are in Fig.~\ref{fig:ablation}.  

\noindent \textbf{Semantic Emotion Guidance (SEG).}  
Removing SEG leads to a consistent degradation across all evaluation metrics, underscoring its crucial role in enabling emotion-aware audio-motion mapping.
A deeper look into its components reveals that removing $\mathcal{L}_{\text{Score}}$ primarily harms audio-motion synchronization due to unstable motion representation, whereas removing $\mathcal{L}_{\text{KL}}$ results in more significant decline in both motion accuracy and visual fidelity, since the residual branch loses its semantic guidance for modeling emotion-driven deformation patterns. 

\noindent \textbf{Expert Motion Decoder.}  
Removing the gate branch introduces temporal instability and weakens audio-motion synchronization.
Disabling the residual branch yields oversmoothed and less expressive motion due to the loss of fine-grained emotional variation.
Eliminating AdaIN-based identity modulation causes the largest performance drop, indicating that it is crucial for preserving identity-specific motion and stabilizing pretraining, without which learning becomes substantially more entangled across identities.

% Ablation Study Table
\begin{table}[t]
\centering
\small
\resizebox{\linewidth}{!}{
\setlength{\tabcolsep}{4pt}
\begin{tabular}{l|cccc}
\toprule
\textbf{Variant} & PSNR$\uparrow$ & LPIPS$\downarrow$ & LMD$\downarrow$ & Sync-C$\uparrow$ \\
\midrule
\textbf{EmoTaG (Full Model)}        &  \textbf{29.95} &  \textbf{0.022} &  \textbf{2.456} &  \textbf{6.147} \\
\midrule
\quad w/o Score Distill ($\mathcal{L}_{\text{Score}}$)  & 29.52 & 0.026 & 2.731 & 5.874 \\
\quad w/o KL Distill ($\mathcal{L}_{\text{KL}}$)        & 29.36 & 0.031 & 2.985 & 5.712 \\
\quad w/o SEG                    & 29.01 & 0.034 & 3.067 & 5.541 \\
\midrule
\quad w/o Gate Branch                                    & 28.77 & 0.036 & 3.358 & 5.004 \\
\quad w/o Residual Branch                                & 28.52 & 0.038 & 3.572 & 4.896 \\
\quad w/o Identity Modulation (AdaIN)                   & 28.38 & 0.040 & 4.021 & 4.621 \\
\bottomrule
\end{tabular}
}
\vspace{-5pt}
\caption{
\textbf{Ablation study of EmoTaG.} 
Evaluated under the 5s \textit{self-reconstruction} setting on the emotional test set, showing the effect of each component on quality and synchronization.
}
\vspace{-6pt}
\label{tab:ablation}
\end{table}

% Ablation Study Figure
\section{Conclusion}
We propose EmoTaG, a novel emotion-aware 3D talking head framework with few-shot personalization. Existing methods struggle on emotional scenarios due to entangled motion representations that overlook prosodic cues and compromise structural stability. EmoTaG addresses this through explicit disentanglement: a Gated Residual Motion Network (GRMN) separates phonetic articulation from emotion-driven modulation, guided by Semantic Emotion Guidance via knowledge distillation. The AdaIN-based modulation further enables efficient adaptation by freezing GRMN and fine-tuning only identity-specific AdaIN parameters from a 5-second video. Comprehensive experiments show that EmoTaG achieves state-of-the-art performance, generating geometrically stable and emotionally coherent talking heads under diverse emotion conditions. Its emotion-aware design effectively bridges physical accuracy and affective expressiveness, enabling more emotionally adaptive 3D talking head avatars.

{
    \small
    \bibliographystyle{ieeenat_fullname}
    \bibliography{main}
}

% \input{sec/6_suppl}

\end{document}